# Utilizing Language Relatedness to improve Machine Translation: A Case Study on Languages of the Indian Subcontinent

Anoop Kunchukuttan*
Microsoft India
ankunchu@microsoft.com

Pushpak Bhattacharyya
Indian Institute of Technology Bombay
pb@cse.iitb.ac.in

**Abstract**

In this work, we present an extensive study of statistical machine translation involving languages of the Indian subcontinent. These languages are related by genetic and contact relationships. We describe the similarities between Indic languages arising from these relationships. We explore how lexical and orthographic similarity among these languages can be utilized to improve translation quality between Indic languages when limited parallel corpora is available. We also explore how structural correspondence between Indic languages can be utilized to re-use linguistic resources for English to Indic language translation. Our observations span 90 language pairs from 9 Indic languages and English. To the best of our knowledge, this is the first large-scale study specifically devoted to utilizing language relatedness to improve translation between related languages.

## 1 Introduction

The term, *related languages*, refers to languages that exhibit lexical and structural similarities on account of sharing a **common ancestry** or being in **contact for a long period of time** (Bhattacharyya et al., 2016). Examples of languages related by common ancestry are Slavic and Indo-Aryan languages. Prolonged contact leads to convergence of linguistic properties even if the languages are not related by ancestry and could lead to the formation of *linguistic areas* (Thomason, 2000). Examples of such linguistic areas are the Indian subcontinent (Emeneau, 1956), Balkan (Trubetzkoy, 1928) and Standard Average European (Haspelmath, 2001) linguistic areas. Genetic as well as contact relationships lead to related languages sharing vocabulary and structural features.

Translation between related languages is an important requirement due to substantial government, business and social communication among people speaking these languages. Another important usecase is translation between a *link language* like English and a set of related languages. However, many of these related languages have few parallel corpora resources, an important requirement for building good quality statistical machine translation (SMT) systems. It is therefore important to utilize the relatedness of these languages to build SMT systems.

Relatedness between these languages leads to a few similarities that are relevant to SMT: (a) lexical similarity, (b) structural correspondence, and (c) morphological isomorphism. *Lexical similarity* means that the languages share many words with the similar form (spelling/ pronunciation) and meaning *e.g. blindness* is represented by the word अन्धापन (*andhapana*) in Hindi and आन्धळेपणा (*aandhaLepaNaa*) in Marathi (both are Indo-Aryan languages). These lexically similar words could be cognates, lateral borrowings or loan words from other languages. *Structural correspondence* means that languages have the same basic word order *viz.* SOV (Subject-Object-Verb), SVO (Subject-Verb-Object), *etc.* The basic word order also determines

*This work was done when the author was a student at IIT Bombay

other syntactic properties like the relative order of noun-adposition, noun-relative clause, noun-genitive, verb-auxiliary, *etc*. Related languages typically tend to possess structural correspondence. *Morphological isomorphism* refers to one-one correspondence between inflectional affixes. While content words are borrowed or inherited across related languages, function words are generally not lexically similar across languages. However, function words in related languages (whether suffixes or free words) tend to have a one-one correspondence to a varying degrees. This may lead to similarities in the case-marking systems of related languages.

In this work, we present an extensive case-study of SMT between related languages of the Indian subcontinent. We focus on studying how relatedness among Indic languages can be used to improve translation quality of SMT systems. The case-study covers 9 Indic languages and 90 language pairs (72 inter-Indic language pairs, 9 English→Indic language and 9 Indic→English language translation pairs).

The Indian subcontinent is home to a large set of related languages. India is one of the most linguistically diverse countries of the world. According to the Census of India of 2001, India has 122 major languages and 1599 other languages. These languages span four major language families. According to *Ethnologue*[1], India has a high Greenberg linguistic diversity index of 0.914 (ranked $14^{th}$ in the world, the highest outside Africa and the Pacific countries of Papua New Guinea, Solomon Islands and Vanuatu). The Indian subcontinent is also home to some of the most widely spoken languages in the world. According to *Ethnologue*[2], seven Indic languages are amongst the top 20 spoken languages in the world: Hindi ($5^{th}$), Bengali ($6^{th}$), Punjabi ($10^{th}$), Telugu ($15^{th}$), Marathi ($16^{th}$), Urdu ($18^{th}$) and Tamil ($20^{th}$). More than 30 languages have more than a million speakers. In addition, English is also widely spoken in India by around 125 million people, though it is not the native language of most speakers.

This paper is organized as follows. Section 2 presents an overview of languages spoken in the Indian subcontinent and their relatedness. Section 3 discussed literature related to our work. Section 4 describes the dataset used in this study. Section 5 presents a study of translation between Indic languages which utilizes lexical similarity between these languages. Section 6 presents a study of translation from English to Indic languages, which utilizes the structural correspondence between Indic languages. Section 7 presents a study of translation from Indic languages to English. Section 8 presents the conclusions from the study and points to possible future directions of work.

## 2 An Overview of Languages of the Indian Subcontinent

This section provides a brief summary about the languages of the Indian subcontinent and their relatedness.

### 2.1 Language Families

India has four major language families, which are briefly summarized in this section.

**Indo-Aryan** It is a sub-family of the larger Indo-European language family. These languages are mainly spoken in North and Central India, and the neighbouring countries of Pakistan, Nepal and Bangladesh. The nearby island countries of Sri Lanka and Maldives also speak Indo-Aryan languages (Sinhala and Dhivehi respectively). The speakers of these languages constitute around 75% of the Indian population.

**Dravidian** It is a language family whose speakers are predominantly found in South India, with some speakers in Sri Lanka and a few minuscule pockets of speakers in North India. Comparative linguistics has not established any conclusive links of Dravidian languages to languages outside India, so these languages could be indigenous to the subcontinent. The speakers of these languages constitute around 20% of the Indian population.

[1] `https://www.ethnologue.com/statistics/country`

[2] `https://www.ethnologue.com/statistics/size`

**Austro-Asiatic** This language family is said to be indigenous to the subcontinent. Languages from the Munda sub-family of the Austro-Asiatic family (the other being Mon-Khmer spoken in South-East Asia) are spoken in the subcontinent, primarily in parts of Central India. Khasi, a non-Munda Austro-Asiatic language, is spoken in some parts North-East India. The languages of the Nicobar islands are also Austro-Asiatic. The speakers of these languages constitute around 5% of the Indian population. The major languages of this group are Santhali and Mundari.

**Sino-Tibetan** Many languages from the Sino-Tibetan language family are spoken in regions of India that border Tibet and South-East Asia, along the Himalayan foothills and North-East India. Most of these languages have a small number of speakers, and these are spoken in areas that are at the intersection of India, China and South-East Asia. The major languages of this group are Meitei and Bodo.

In addition, an endangered set of languages from the Great Andamanese language family is spoken on the Andaman islands. Moreover, no information is available on the language(s) of the Sentinelese people of the Andaman, who have no contact with the world outside their native Sentinel island.

## 2.2 Writing Systems

The major Indic languages have a long written tradition and use a variety of scripts. These scripts are derived from the ancient *Brahmi* script. These are *abugida* scripts where the organizing unit is the *akshar*, a consonant cluster along with an optional *matra* (vowel diacritic). All these scripts have a high grapheme to phoneme correspondence and represent almost the same set of phonemes. This similarity is useful for utilizing orthographic and phonetic similarities across these languages. However, the visual layout of the characters is very different across languages; hence, each script has its own designated range of codepoints in the Unicode standard. Some of the major languages using Brahmi-derived scripts are Sanskrit, Hindi, Bengali, Tamil and Telugu.

However, there are many languages that do not use Brahmi-derived scripts. Prominent among these is Urdu, which uses an Arabic-derived script. Kashmiri, Punjabi and Sindhi use Brahmi-derived as well as Arabic-derived scripts. Many languages in Central India and North-East India, which did not have a literary tradition in the past, have adopted the Latin script or one of the various Brahmi-derived scripts in modern times.

## 2.3 Relatedness among Indic Languages

Underlying the vast diversity in Indic languages are many commonalities. The languages within each language family are obviously related. In addition, because of contact over thousands of years, the linguistic features of languages belonging to different language families have also undergone convergence to a large extent. Hence, linguists typically refer to India as a *linguistic area* (Emeneau, 1956). In this section, we describe the relatedness among various Indic languages.

### 2.3.1 Lexical Similarity

Languages in the same language family obviously share many cognates. Figure 1 shows examples of cognates from Indo-Aryan languages. In addition, there are many borrowed words between the language families. Many Dravidian languages have borrowed a lot of words from Sanskrit, an Indo-Aryan language; some examples are shown in Table 2. Indo-Aryan languages have also borrowed words from Dravidian languages.

Table 1: Examples of cognates in Indo-Aryan languages

| Hindi | Gujarati | Marathi | Bengali | Meaning |
|---|---|---|---|---|
| रोटी (*roTI*) | રોટલો (*roTalo*) | चपाती (*chapAtI*) | রুটি (*ruTi*) | bread |
| मछली (*maChlI*) | માછલી (*mAChlI*) | मासा (*mAsa*) | মাছ (*mACha*) | fish |
| भाषा (*bhAShA*) | ભાષા (*bhAShA*) | भाषा (*bhAShA*) | ভাষা (*bhAShA*) | language |
| दस (*dasa*) | દસ (*dasa*) | दहा (*dahA*) | দশ (*dasha*) | ten |

Table 2: Examples of some words borrowed from Sanskrit into Dravidian languages

| Sanskrit Word | Dravidian Language Word | Dravidian Language | Meaning |
|---|---|---|---|
| चक्रम् (*cakram*) | சக்கரம் (*cakkaram*) | Tamil | wheel |
| मत्स्यः (*matsyaH*) | మత్స్యలు (*matsyalu*) | Telugu | fish |
| अश्वः (*ashvaH*) | ಅಶ್ವ (*ashva*) | Kannada | horse |
| जलम् (*jalam*) | ജലം (*jala.m*) | Malayalam | water |

*e.g.* the word for *fruit* in Tamil is பழம் (*pazha.m*). It has been borrowed into Indo-Aryan languages; in Hindi, the word is फल (*phala*). Even fixed expressions like idioms which are culture-specific have been borrowed across languages. *e.g.* The Hindi idiom दाल में कुछ काला होना (*dAla me.n kuCha kAlA honA*) and the Gujarati idiom દાળ મા કાઈક કાળુ હોવુ (*dALa mA kAIka kALu hovu*) are very similar lexically and essentially mean the same - *something is suspicious* though it literally translates to *something is black in the lentils*.

We attempt to quantify the lexical similarity between Indic languages using a simple metric to provide an indicative measure of lexical similarity. We use the Longest Common Subsequence Ratio (LCSR) (Melamed, 1995) as a measure of the lexical similarity[3]. For a pair of languages, we compute the average LCSR between every pair of sentences in the training corpus at the character level. This is an obvious approach since lexical similarity stems from subword-level correspondences and the word order is roughly the same between related languages. In order to compare text across different scripts, we map all the Indic scripts to the Devanagari script. We used the n-way ILCI parallel corpus for this study (corpus details are described in Section 4).

Figure 1 shows the lexical similarities of all the language pairs used in our experiments. The lexical similarities reflect the common understanding of similarity between Indic languages. For instance,

- We see that Hindi is closer to Punjabi than to Marathi.
- Marathi and Konkani are most similar to each other.
- The Dravidian languages Malayalam, Telugu and Tamil are closer to each other, than to Indo-Aryan languages.
- Dravidian and Indo-Aryan languages also show a reasonable level of lexical similarity between them due to contact between these language families over a long time.
- Telugu has higher similarity with Indo-Aryan languages than other Dravidian languages; Telugu speakers border Indo-Aryan speakers and hence exhibit greater lexical convergence.

Note that the lexical similarities values involving Tamil should be read with caution since the Tamil script is underspecified. The Tamil script does not have characters to represent voiced as well as aspirated plosives; these are represented by the corresponding unvoiced, unaspirated plosive.

[3]We found a high correlation between LCSR values and manually judged lexical similarity values for a few European languages. These manual judgements were obtained from *Ethnologue* for some European language pairs, but are not available for Indic languages. Nevertheless, this result demonstrates that LCSR is a reliable metric to quantify lexical similarity automatically.

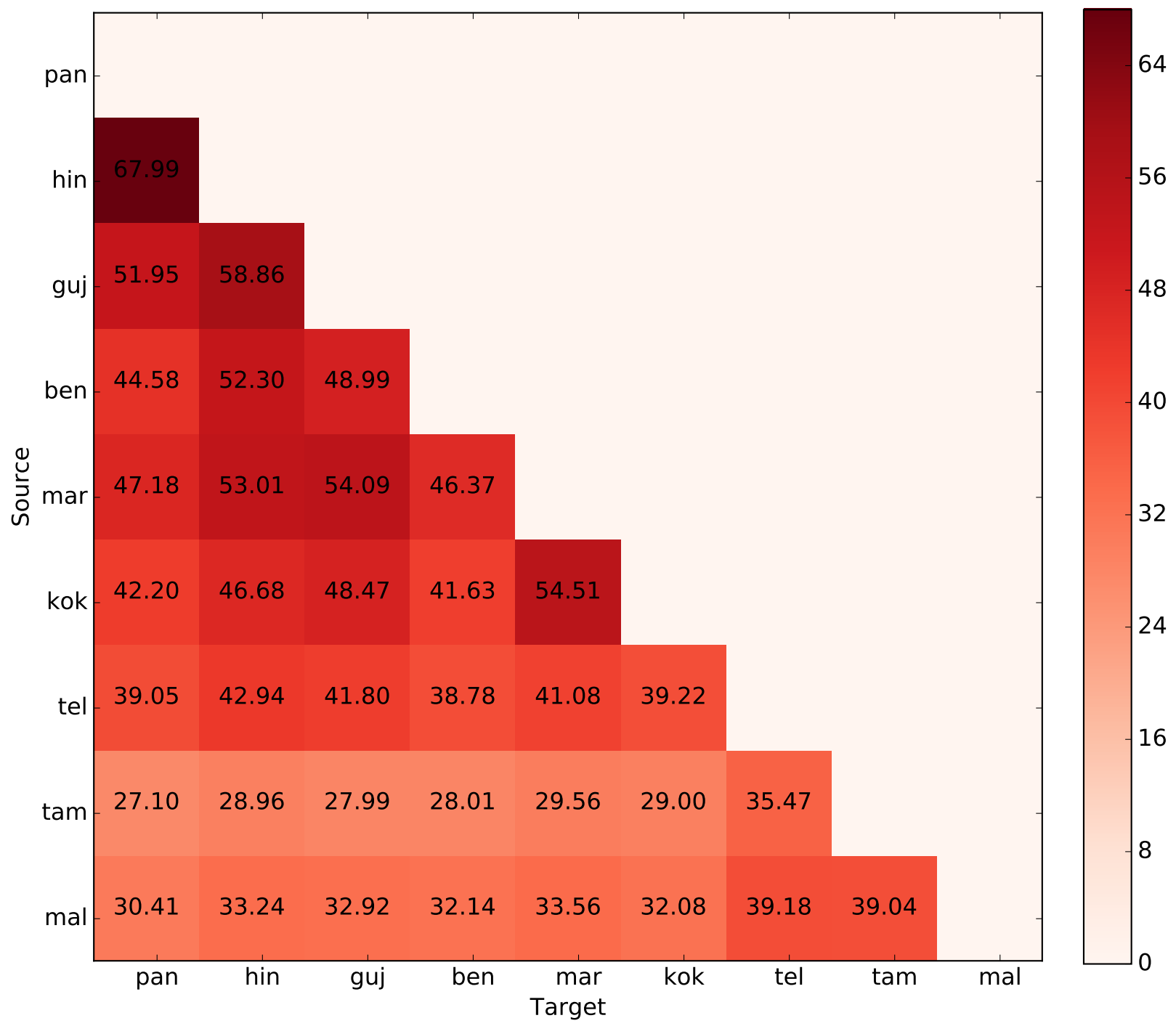

Figure 1: Lexical Similarity between major Indic Languages

### 2.3.2 Structural Correspondence

Languages of all the four major language families have the same word order: Subject-Object-Verb. In fact, the Munda languages were originally supposed to be SVO languages like their South-East Asian Mon-Khmer relatives; but, they have transformed to SOV languages (Subbārāo, 2012). The only exceptions are Khasi, Nicobarese and Kashmiri (which are all SVO languages) (Abbi, 2012).

### 2.3.3 Other Examples of Linguistic Convergence

The following are a few examples of borrowing of language features across language families to illustrate the extent of convergence between Indic languages.

**Retroflex Sounds** (Emeneau, 1956; Abbi, 2012): These sounds are present in the Indo-Aryan languages, but not found in Indo-European languages outside the subcontinent. They were borrowed into Indo-Aryan languages from either the Dravidian or Austro-Asiatic languages, which possess these sounds. In general, there is a high degree of overlap between the phoneme set of Indic languages.

**Echo Words** (Emeneau, 1956; Subbārāo, 2012): Again, this feature is unique to Indo-Aryan languages amongst the Indo-European languages. They are a standard feature of Dravidian languages.

**Dative Subjects** (Abbi, 2012): The dative subject represents a non-agentive subject, generally the experiencer. The subject is marked in the dative case, whereas the direct object is marked with the nominative case.

**Conjunctive Particles** (Subbārāo, 2012; Abbi, 2012): These are used to conjoin two verb phrases in a manner similar to conjunction. The two verb phrases represent two sequential actions; first action expressed with a conjunctive participle.

**Quotative Verb** (Subbārāo, 2012; Abbi, 2012): It reports someone else's quoted speech. This feature is present in Dravidian, Munda, Tibeto-Burman and some Indo-Aryan languages.

**Compounds Verb** (Subbārāo, 2012; Abbi, 2012): It refers to verbs composed of a primary verb followed by a vector verb. The primary verb carries the semantics, whereas the vector verb is limited to a finite set of words and marks certain grammatical properties of the main verb.

**Conjunct Verb** (Subbārāo, 2012): It refers to a noun+verb combination, where the semantics is carried by the noun while the 'light' verb carries various grammatical markers.
Given the high level of convergence, it has been said that (Abbi, 2012):

> *"India as a linguistic area gives us robust reasons for writing a common or core grammar of many of the languages in contact."*

# 3 Related Work

We first summarize past work on utilizing language relatedness for machine translation. Then, we look at large-scale studies on translation for related languages, including Indic languages.

## 3.1 Utilizing Language Relatedness

Lexical similarity has been used to improve translation between related languages via sub-word level transformations to translate cognates and borrowed words. One approach involves *transliteration of source words* into the target language. The transliteration candidates can compete with translation candidates during decoding (Durrani et al., 2010) or transliteration can be a post-decoding step applied to untranslated words (Nakov and Tiedemann, 2012; Kunchukuttan et al., 2014b).

Since a high degree of similarity exists at the subword-level between related languages, the second approach looks at *translation with subword level basic units*. Character-level SMT has been explored for very closely related languages like *Bulgarian-Macedonian, Indonesian-Malay, Spanish-Catalan* with modest success (Vilar et al., 2007; Tiedemann, 2009; Tiedemann and Nakov, 2013). Character n-gram units have been used to provide a better contextual representation, but provides little benefit beyond $n = 2$ since vocabulary size increases for higher order n-grams (Tiedemann and Nakov, 2013).

Some variable length units provide larger subword units while controlling the vocabulary size. *Orthographic syllables* (Kunchukuttan and Bhattacharyya, 2016b) are approximate syllables that have shown good performance even when the related languages are not very close or even in the same language family. Statistically discovered frequent subwords like *Byte-pair encoded units* have also shown good performance and shown to balance utilization of lexical similarity with word-level information (Kunchukuttan and Bhattacharyya, 2017). Such subwords were first proposed in the context of Neural MT: BPE (Sennrich et al., 2016), wordpieces (Schuster and Nakajima, 2012; Wu et al., 2016), Huffman encodings (Chitnis and DeNero, 2015).

Syntactic similarity has also been utilized for reducing resource requirements. Source-side pre-ordering is a useful method to address word order divergence between source and target languages (Collins et al., 2005). It has been shown that source reordering rules written for one language can be successfully re-used for another related language (Kunchukuttan et al., 2014a).

Sometimes, a direct parallel corpus is not available between the related languages, but they share a parallel corpus with a pivot language. In that case, pivot-based approaches can be used to learn word, phrase and/or morpheme level mappings between the related languages. The sentence translations can be generated by searching though these potential mappings. The similar word order of the two languages can be used to simplify this search. This approach has been explored in some works (Wang et al., 2016).

## 3.2 Large-scale Studies on Related Languages

We summarize previous case-studies involving machine translation for related languages. We cover Indic and non-Indic languages separately.

### 3.2.1 Non-Indic languages

A couple of works describe and analyze translation systems for all language pairs in the *Europarl* and *Acquis Communautaire* corpus respectively (Koehn, 2005; Koehn et al., 2009). The former work spans 11 languages and builds 110 translation systems. The latter work spans 22 languages and builds 462 translation systems. The languages covered are all European languages. They build word-level PBSMT systems for these languages pairs. Both the works report that language relatedness affects translation quality, but do not propose any solution to utilize language relatedness. Both the works also suggest that rich morphology is a challenge to building machine translation systems.

A previous work has created the *South-East European Times Parallel Corpus* of Balkan languages and trained word-level PBSMT systems for 72 language pairs (Tyers and Alperen, 2010). While the authors envision this as a step towards pan-Balkan translation, the work does not study the effect of language relatedness or its utilization for improvement of translation. On a smaller scale, the Asian Language Treebank containing small parallel corpora for many South-East Asian languages has been developed (Thu et al., 2016). Baseline PBSMT system scores for two language pairs from this corpus have been reported (Ding et al., 2016).

### 3.2.2 Indic languages

English to multiple Indic language phrase-based SMT systems have been explored by (Post et al., 2012) (6 Indic languages from crowd-generated corpora) and the *Anuvadaksh* project (8 Indic languages)[4]. No evaluation of the latter system is available in the public domain, except for the English-Hindi SMT engine (Ramanathan et al., 2008; Patel et al., 2013).

There has been a large-scale study of Indic language to Indic language SMT systems, and English to Indic language SMT systems for 110 language pairs (Kunchukuttan et al., 2014a). This work showed that language relatedness impacts translation and there is a clear partitioning among Indo-Aryan and Dravidian languages with respect to translation quality. Another work compared NMT and SMT systems for 110 Indic language pairs (Agrawal, 2017). They have focussed on improving the NMT systems using linguistic features and monolingual corpora. They found that NMT systems made fewer morphological and syntax/agreement errors, but more lexical choice errors compared to SMT systems. Both the works train word-level models and have not investigated how subword level models can be used to improve NMT systems.

A lot of the previous work in pan-Indic language MT has involved rule-based MT systems. The *AnglaBharati* system (Sinha et al., 1995) is an *English-to-Indic* language based pseudo interlingua-based MT system which harnesses the common characteristics of Indic languages in the syntax transfer stage. The syntax transfer change is common to all languages and generates a pseudo-target language output corresponding to

[4] `http://tdil-dc.in/index.php?option=com_vertical&parentid=72`

Table 3: Monolingual corpora used for building word-level language models

| Language | Sources | Sentences |
|---|---|---|
| hin | HindMonoCorp (Bojar et al., 2014) | 10,044,777 |
| pan | Leipzig Corpus (Quasthoff et al., 2006) | 144,777 |
| guj | Leipzig Corpus (Quasthoff et al., 2006) | 444,777 |
| ben | Leipzig Corpus (Quasthoff et al., 2006) | 436,689 |
| mar | Leipzig Corpus (Quasthoff et al., 2006), News Websites | 1,923,688 |
| kok | Leipzig Corpus (Quasthoff et al., 2006) | 84,777 |
| tel | Leipzig Corpus (Quasthoff et al., 2006) | 644,777 |
| tam | Leipzig Corpus (Ramasamy et al., 2012) | 1,484,777 |
| mal | Leipzig Corpus (Quasthoff et al., 2006) | 184,777 |

Indic language word order. The *Sampark* system[5] (Anthes, 2010) is a transfer-based system for translation between 9 Indic language pairs that uses a common lexical transfer engine, whereas minimum structural transfer is required between Indic languages. The emphasis is on detailed morphological analysis to enable accurate lexical transfer and target generation. Lexical similarity has not be harnessed in any significant measure in these systems.

# 4 Datasets

We used the Indian Language Corpora Initiative (ILCI) corpus[6] (Jha, 2012) for our experiments. The ILCI corpus is an 11-way multilingual corpus of sentences from the health and tourism domains. The Indic languages we experimented with are:

- **Indo-Aryan**: Hindi, Punjabi, Bengali, Gujarati, Marathi, Konkani
- **Dravidian**: Telugu, Tamil, Malayalam

These languages represent the two major language families in India. We also experiment with English to Indic language translation and *vice versa*. The data split is as follows (in number of sentences): **Train**: 44,777, **Tune**: 1,000, **Test**: 2000

For training word-level language models, we used monolingual corpora from different sources in addition to the train split from the parallel corpora. The details are provided in Table 3.

# 5 Translation between Indic Languages

In this study on translation between related Indic languages, we compare translation using meaning-bearing units (words and morphemes) with translation methods which can utilize lexical similarity : (a) subword-level translation models (with OS and BPE units), (b) transliteration of untranslated words from the output of a word-level SMT system. We train translation systems between 72 Indic language pairs.

## 5.1 Subword-level Translation Models

We explore *orthographic syllable (OS)* and *Byte Pair Encoded unit (BPE)* as subword units.

[5]http://sampark.iiit.ac.in

[6]The corpus is available on request from http://tdil-dc.in

Both OS and BPE units are variable length units which provide appropriate context for translation between related languages. Since their vocabularies are much smaller than the morpheme and word-level models, data sparsity is not a problem. OS and BPE units have outperformed character n-gram, word and morpheme-level models for SMT between related languages (Kunchukuttan and Bhattacharyya, 2016b, 2017).

While OS units are approximate syllables, BPE units are highly frequent character sequences, some of them representing different linguistic units like syllables, morphemes and affixes. While orthographic syllabification applies to writing systems which represent vowels (alphabets and abugidas), BPE can be applied to any script.

#### 5.1.1 Orthographic Syllable

The *orthographic syllable*, a **linguistically motivated unit**, is a sequence of one or more consonants followed by a vowel, *i.e.,* a $C^+V$ unit (*e.g. spacious* would be segmented as *spa ciou s*). Note that the vowel character sequence *iou* represents a single vowel. It represents an approximate syllable, with onset and nuclues, but no coda. Orthographic syllabification is rule-based and applies to writing systems which represent vowels (alphabets and abugidas).

#### 5.1.2 Byte Pair Encoded Units

The *BPE unit* is motivated by **statistical properties of text** and represents stable, frequent character sequences in the text (possibly linguistic units like syllables, morphemes, affixes). Given monolingual corpora, BPE units can be learnt using the Byte Pair Encoding text compression algorithm (Gage, 1994). The BPE algorithm is an iterative one that discovers frequent substrings in the text, starting with bigram substrings and discovering longer substrings (Sennrich et al., 2016). BPE can be applied to text in any writing system.

#### 5.1.3 Training Considerations

We segment the data into subwords during pre-processing and indicate word boundaries by a boundary marker (_) as shown in the example below. The boundary marker helps keep track of word boundaries, so the word level representation can be reconstructed after decoding.

word: `Childhood means simplicity .`

subword: `Chi ldhoo d _ mea ns _ si mpli ci ty _ .`

While building phrase-based SMT models at the subword level, we use (a) monotonic decoding since related languages have similar word order, (b) higher order languages models (10-gram) since data sparsity is a lesser concern owing to small vocabulary size (Vilar et al., 2007), and (c) word level tuning (by post-processing the decoder output during tuning) to optimize the correct translation metric (Nakov and Tiedemann, 2012). Following decoding, we used a simple method to regenerate words from subwords (de-segmentation): concatenate subwords between consecutive occurrences of boundary marker characters.

### 5.2 Transliteration of Untranslated Words

We transliterate the untranslated words from the output of a word-level PBSMT system. Since the source and target languages are lexically similar, we expect that the transliteration would result in successful translation of many untranslated words. We explore two transliteration methods:

**Rule-based Transliteration** The rule-based transliteration relies on the orthographic similarity between Indic scripts. There is an almost one-one correspondence between scripts of major Indic languages, which derive from the Brahmi script. Further, the design of the Unicode standard utilizes the orthographic similarity and makes rule-based transliteration very trivial. The Unicode standard assigns blocks of 128 codepoints each to various Indic scripts (*e.g.* U+0900-U+097F for Devanagari, U+0D00-U+0D7F for Malayalam). The first 112 characters, covering the major characters, are aligned across all scripts by placing them at a common offset with respect to the start of the Unicode range for that script. For instance, the Devanagari character क (*ka*, U+0915) and the corresponding Malayalam character ക (*ka*, U+0D15) are both at offset 16 within their respective Unicode ranges. This makes transliteration simply a matter of manipulating the start of the Unicode ranges. The transliteration rule is illustrated below:

$$char_{tgt} = \texttt{to_char}(\texttt{to_codept}(char_{src}) - \texttt{rstart}(lang_{src}) + \texttt{rstart}(lang_{tgt})) \tag{1}$$

where,
$lang_{src}$ and $lang_{tgt}$ are source and target languages respectively
$char_{src}$ and $char_{tgt}$ are source and target characters respectively
`to_codept` is a function that returns the Unicode code point (value) corresponding to the Unicode character supplied as argument
`to_char` is a function that returns the Unicode character corresponding to the Unicode code point (value) supplied as argument
`rstart` is a function that returns the codepoint (value) corresponding to the beginning of the Unicode range for the language supplied as argument.

The single rule above is sufficient for script conversion among many Indic languages. The orthographic similarity between languages was utilized in the design of the Unicode standard, which in turn enabled sharing of the script conversion rule among many Indic languages.

**Statistical Transliteration** We use the *BrahmiNet* transliteration system (Kunchukuttan et al., 2015) for statistical transliteration. This system uses a PBSMT model to learn transliteration systems between Indic languages. It uses a parallel transliteration corpus mined from the ILCI parallel translation corpus using the Transliteration Module in Moses (Sajjad et al., 2012; Durrani et al., 2014). Since the transliteration pairs were mined from the translation parallel corpus, the mined pairs are representative of diverse lexical similarity phenomena — spelling variations, sound shifts, cognates and loan words. Hence, it is appropriate for training transliteration systems to transliterate untranslated words.

## 5.3 Experimental Settings

We trained phrase-based SMT systems using the *Moses* system (Koehn et al., 2007), with the *grow-diag-final-and* heuristic for extracting phrases, and Batch MIRA (Cherry and Foster, 2012) for tuning (default parameters). We trained 5-gram LMs with Kneser-Ney smoothing for word and morpheme level models and 10-gram LMs for OS and BPE-unit level models. Subword level representation of sentences is long, hence we speed up decoding by using cube pruning with a smaller beam size (pop-limit=1000). This setting has been shown to have minimal impact on translation quality (Kunchukuttan and Bhattacharyya, 2016a).

### 5.3.1 Word Segmentation

We used unsupervised morphological-segmenters for generating morpheme representations (trained using *Morfessor* (Smit et al., 2014)). We used the models distributed as part of the *Indic NLP Library*[7] (Kunchukuttan et al., 2014b). We used orthographic syllabification rules from the *Indic NLP Library* for Indic languages. For training BPE models, we used the *subword-nmt*[8] library.

[7]http://anoopkunchukuttan.github.io/indic_nlp_library
[8]https://github.com/rsennrich/subword-nmt

### 5.3.2 Evaluation

We evaluate all our models using *word-level* BLEU (Papineni et al., 2002). Note that though we train models with various kinds of subwords, the word-level output is generated before BLEU evaluation. Hence, the reported scores are comparable.

## 5.4 Results

### 5.4.1 Word-level and Morpheme-level Translation

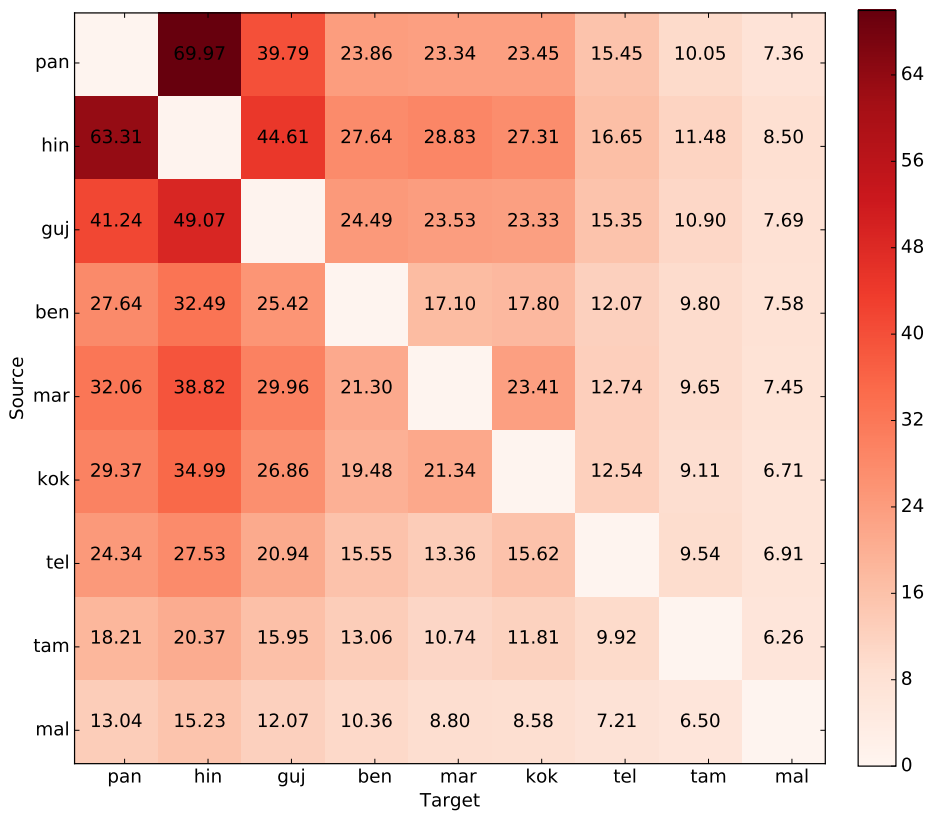

(a) Translation quality for word-level models (BLEU)

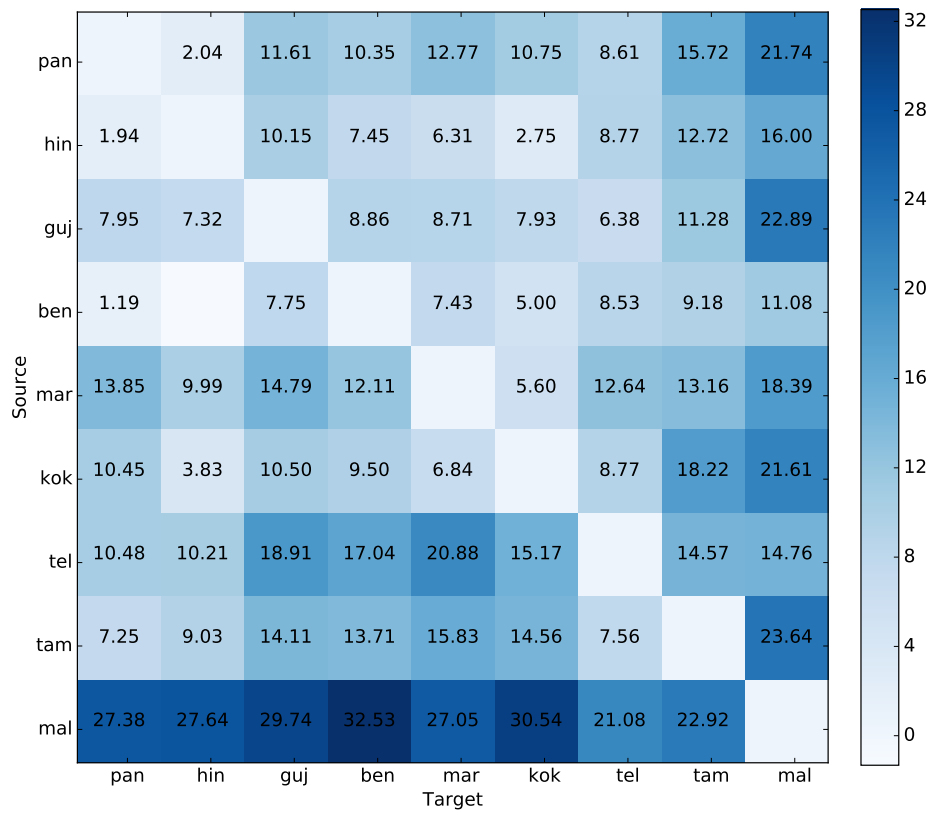

(b) % difference w.r.t. word-level for morpheme-level models

Figure 2: Word and Morpheme-level Translation

The baseline word and morpheme-level models were trained as described earlier. They do not utilize the lexical similarity between languages.

**Word-level Translation** Figure 2a shows translation quality for word-level translation (BLEU). We see that there is a clear partitioning of the BLEU scores as per language families. Translation between Indo-Aryan languages is the easiest, while translation between Dravidian languages is the most difficult. Moreover, translation into Dravidian languages is more difficult compared to translating from Dravidian languages. The agglutinative nature of the Dravidian languages plays an important role in making translation involving Dravidian languages challenging since it leads to data sparsity and untranslated words.

**Morpheme-level Translation** Morpheme-level translation models are a way to address data sparsity. Figure 2b shows the % change in BLEU scores compared to word-level models. We observe an average increase of 12.9% in BLEU score compared to word-level models. We see a larger increase for translation involving Dravidian languages. Translation between Dravidian languages shows an improvement of 17.4%, while translation between Indo-Aryan languages shows an improvement of 7.8%.

### 5.4.2 Subword-level Translation

While morpheme-level models reduce data sparsity and utilize morphological isomorphism between source and target languages, they cannot utilize lexical similarity between languages. To utilize lexical similarity, we train OS-level and BPE-level models (with the number of BPE merge operations=1000), using the same procedure as described earlier.

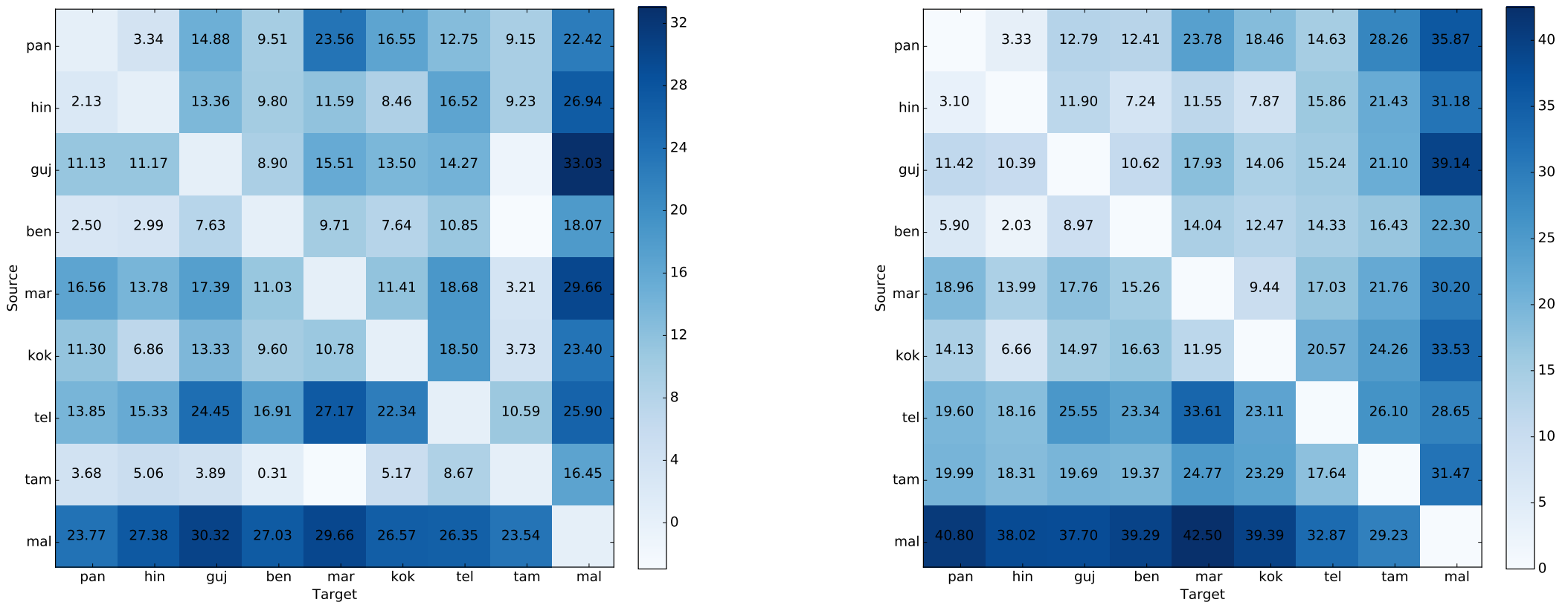

(a) % difference w.r.t. word-level for OS-level (b) % difference w.r.t. word-level for BPE-level

Figure 3: Difference in translation quality (BLEU) between subword-level models and word-level models

**OS-level models** Figure 3a shows the % change in BLEU scores of OS-level models compared to word-level models. We observe an average increase of 20.2% in BLEU score compared to word-level models. Again, we see a larger increase for translation involving Dravidian languages (27.7% improvement).

**BPE-level models** Figure 3b shows the % change in BLEU scores of BPE-level models compared to word-level models. We observe an average increase of 20.2% in BLEU score compared to word-level models. Again, we see a larger increase for translation involving Dravidian languages (27.7% improvement). The BPE-level models show 6.1% improvement over morpheme-level models.

**Thus, the use of lexical similarity helps improve the translation quality**.

#### 5.4.3 Transliteration of Untranslated Words

Figure 4a shows the change in transliteration scores due to the simple rule-based transliteration scheme described earlier. We see a modest improvement of about 1% in BLEU score after transliteration of the untranslated words using this simple scheme. This is a very simple transliteration scheme that does not take into account phonetic variations and change in spelling conventions across the languages - it can be more appropriately referred to as *script conversion*. Hence, it is not surprising that we do not get a significant improvement in translation quality; even if the transliteration is off by a single character, it would not contribute to improvement in BLEU score. Nevertheless, it improves the perceived translation quality for users since they can read the untranslated words in the target script and guess the named entities and cognates. Note that no parallel transliteration corpus was required for transliteration.

Figure 4b shows the improvement in translation quality due to statistical transliteration of untranslated words. We observe an average improvement of 5.3% in translation quality (BLEU). Note that this improvement is less than the improvement achieved using subword-level translation. This reinforces results from previous studies that subword-level translation is better than the transliteration of untranslated words in the output of an SMT system.

### 5.5 Discussion

#### 5.5.1 Analysis by Language Group

While the previous sections reported differences in translation quality for every language pair, this section looks at average differences in translation quality for translation among different language groups. The two

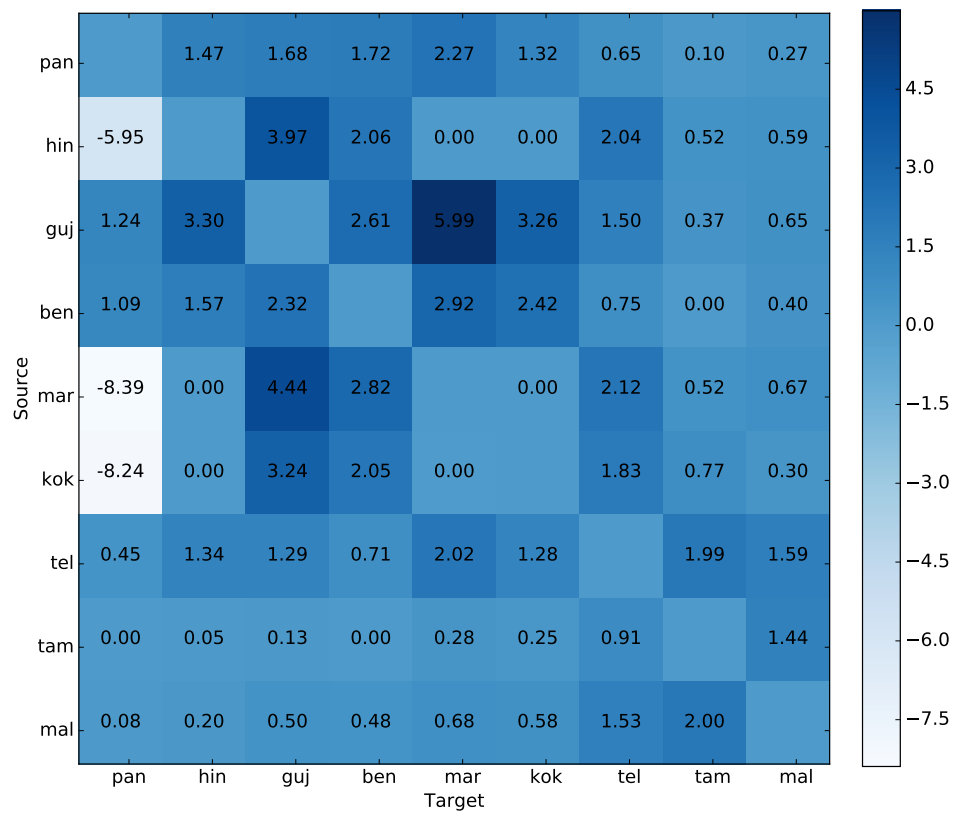

(a) % difference in BLEU w.r.t. word-level for rule-based transliteration of untranslated words

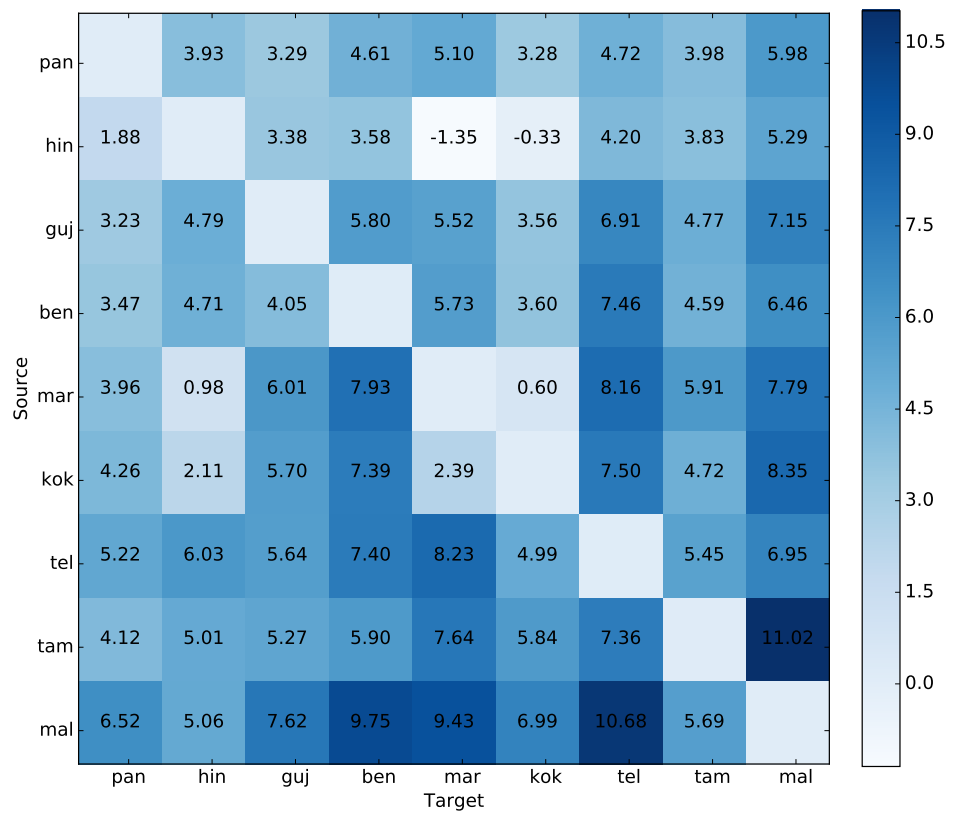

(b) % difference in BLEU w.r.t. word-level for statistical transliteration of untranslated words

Figure 4: Effect of transliteration of untranslated words from a word-level translation model

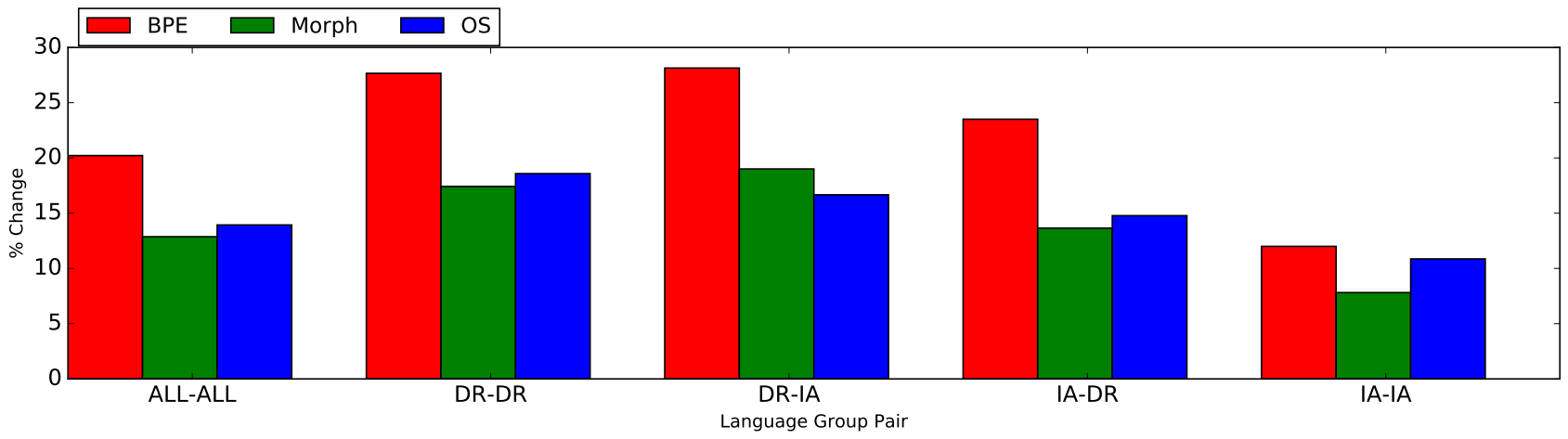

Figure 5: Comparison of average change in translation quality (BLEU) for various language groups and translation units

language groups under analysis are: Indo-Aryan (IA) and Dravidian (DR). Hence, the pairs of language groups for translation are: IA-IA, IA-DR, DR-IA and DR-DR *i.e.,* IA-IA indicates translation between Indo-Aryan languages and so on. Figure 5 shows the average % improvement in BLEU scores for translation between different language groups. ALL-ALL indicates average across all languages, irrespective of the language group. These improvements are reported for BPE, OS and morpheme-level models compared to baseline word-level models. The maximum improvements are seen for Dravidian to Indo-Aryan (DR-IA) and Dravidian to Dravidian (DR-DR) translation. Thus, morphologically rich languages benefit the most from subword-level translation.

### 5.5.2 Easiest/Difficult Languages to Translate

To derive insights into which languages are easy to translate from, we compute the average BLEU scores translating from a language into other languages (and *vice versa*). Table 4 shows these average BLEU scores for BPE-level models. Other translation units also show the same trends. We see that Hindi is the easiest language to translate into/from other languages. On the other hand, Malayalam is the most difficult language to translate into/from other languages. In general, translation involving Indo-Aryan languages is easier and translation involving Dravidian languages is more difficult. Morphological richness is a major distinguishing factor between Indo-Aryan and Dravidian languages and seems to be a challenging problem to address in order to further improve translation involving Dravidian languages. Marathi and Konkani, which are the most morphologically complex Indo-Aryan languages, happen to be the most difficult to translate. It is interesting to note that Bengali seems to be more difficult to translate than many other Indo-Aryan

Table 4: Average BLEU scores translating to/from different languages

(a) Translating from a language

| Source | Ave. BLEU |
|---|---|
| hin | 31.36 |
| pan | 30.16 |
| guj | 27.90 |
| ben | 20.52 |
| mar | 25.53 |
| kok | 22.97 |
| tel | 20.64 |
| tam | 16.06 |
| mal | 14.11 |

(b) Translating into a language

| Target | Ave. BLEU |
|---|---|
| hin | 39.86 |
| pan | 35.18 |
| guj | 31.33 |
| ben | 22.50 |
| mar | 22.00 |
| kok | 21.91 |
| tel | 14.97 |
| tam | 11.87 |
| mal | 9.61 |

languages. The phonetics of Bengali varies to some extent from other Indo-Aryan languages, and Bengali shows influence from Tibeto-Burman languages too. This could be a potential reason for the comparative difficulty in translation involving Bengali.

# 6 Translation from English to Indic Languages

English is a Subject-Verb-Object language, while the canonical word order in all major Indic language is Subject-Object-Verb (though these languages are free word order to a reasonable extent). This is the fundamental divergence that needs to be bridged for translation from English to Indic languages. We first built a baseline PBSMT system, followed by two source-side pre-reordering systems.

## 6.1 Baseline PBSMT

We trained word-level, phrase-based SMT systems for translation from English to Indic languages using *Moses* (Koehn et al., 2007), with the *grow-diag-final-and* heuristic for symmetrization of word alignments and the *msd-bidirectional-fe* model for lexicalized reordering (Tillmann, 2004). We tuned the trained models using Batch MIRA (Cherry and Foster, 2012) with default parameters. We trained 5-gram language models on the target side corpus with the Kneser-Ney smoothing using SRILM (Stolcke et al., 2002).

## 6.2 Reusing Source-side Pre-ordering rules

The following generic transformation principle going from English to Hindi word order can be used (Ramanathan et al., 2008):

$$SS_mVV_mOO_mC_m \quad \leftrightarrow \quad C'_mS'_mS'V'_mV'O'_mO' \tag{2}$$

where,
$S$: Subject, $O$: Object, $V$: Verb, $C_m$: Clause modifier
$X'$: Corresponding constituent in Hindi.
$X$ is S, O or V
$X_m$: modifier of X
The following is an example of the application of the generic rule:

| | |
|---|---|
| English | $\overbrace{\text{the hero}}^{S}\overbrace{\text{of the movie}}^{S_m}\overbrace{\text{shot}}^{V}\overbrace{\text{the scene}}^{O}\overbrace{\text{quickly}}^{V_m}$ |
| English Pre-ordered | $\overbrace{\text{the movie of}}^{S_m}\overbrace{\text{the hero}}^{S}\overbrace{\text{quickly}}^{V_m}\overbrace{\text{the scene}}^{O}\overbrace{\text{shot}}^{V}$ |
| Hindi | $\overbrace{\text{फिल्म के}}^{S_m}\overbrace{\text{नायक ने}}^{S}\overbrace{\text{जल्दी}}^{V_m}\overbrace{\text{दृश्य}}^{O}\overbrace{\text{शूट किया}}^{V}$ |
| Hindi (ITRANS) | $\overbrace{\text{philma ke}}^{S_m}\overbrace{\text{nAyaka ne}}^{S}\overbrace{\text{jaldI}}^{V_m}\overbrace{\text{dRîshya}}^{O}\overbrace{\text{shUTa kiyA}}^{V}$ |

Table 5: English to Indic Language Translation (BLEU)

| Pre-ordering | pan | hin | guj | ben | mar | kok | tel | tam | mal |
|---|---|---|---|---|---|---|---|---|---|
| None | 15.83 | 21.98 | 15.80 | 12.95 | 10.59 | 11.07 | 7.70 | 6.53 | 3.91 |
| Generic | 17.06 | 23.70 | 16.49 | 13.61 | 11.05 | 11.76 | 7.84 | 6.82 | **4.05** |
| Hindi-tuned | **17.96** | **24.45** | **17.38** | **13.99** | **11.77** | **12.37** | **8.16** | **7.08** | 4.02 |

They showed that source-side pre-ordering rules, based on the above principle, improve English-Hindi translation quality. This principle holds across all Indic languages, hence the rules have been shown to work for other Indic languages as well (Kunchukuttan et al., 2014a). We call these the *generic pre-ordering* rules for English-Indic language translation. Further, refinements to these generic rules have been proposed based on an error analysis of English-Hindi translation output (Patel et al., 2013). We call these refined rules the *Hindi-tuned pre-ordering* rules. It has been shown that these Hindi-tuned rules can be successfully applied to other Indic languages also, taking advantage of the structural correspondence between Indic languages. Our results reaffirm these earlier studies.

### 6.3 Results and Discussion

Table 5 shows BLEU scores for various English-Indic language translation systems. The following are the major observations:

- We observe that both that the source-side pre-ordering systems improve translation quality for all the Indic languages, though they had previously been developed and tested for English-Hindi translation. The *generic* system shows a 5% average improvement in BLEU score, whereas the *Hindi-tuned* system shows a 9% improvement in BLEU score.

- The average improvement over all Indic languages roughly corresponds to the improvement in English-Hindi translation, showing that the pre-ordering rules are just as useful for other Indic languages as they are for Hindi.

- The *Hindi-tuned* system, which was customized based on English-Hindi error analysis, is better than the *generic* system for other Indic languages. On an average, the *Hindi-tuned* system improves the BLEU scores by 4% over the *generic* system.

- Both the reordering systems show better improvement for Indo-Aryan languages compared to Dravidian languages. Since there are some syntactic divergences between Indo-Aryan and Dravidian languages, there may be a case for customizing the rules for Dravidian languages.

Table 6: Indic Language to English Translation (BLEU)

| Method | pan | hin | guj | ben | mar | kok | tel | tam | mal |
|---|---|---|---|---|---|---|---|---|---|
| PBSMT | 22.12 | 24.70 | 19.21 | 17.39 | 16.48 | 16.27 | 13.60 | 11.78 | 8.87 |

# 7 Translation from Indic Languages to English

We trained phrase-based SMT systems for translation from Indic languages to English using the *Moses* (Koehn et al., 2007), with the *grow-diag-final-and* heuristic for symmetrization of word alignments and the *msd-bidirectional-fe* model for lexicalized reordering (Tillmann, 2004). We tuned the trained models using Batch MIRA (Cherry and Foster, 2012) with default parameters. We trained 5-gram language models on the target side corpus with the Kneser-Ney smoothing using SRILM (Stolcke et al., 2002).

Table 6 shows the BLEU scores. Obviously, lexicalized reordering is not sufficient for handling structural divergences between Indic languages and English. These results are initial results, which can be improved with source-side pre-reordering and syntax-based MT methods. Currently, these approaches are not feasible since most Indic languages do not have a constituency or dependency parser available. In the spirit of the utilizing language relatedness, parsers for Indic languages must define the same set of dependency relations, and a common parsing framework for Indic languages must be defined. Efforts in this direction are underway by different research groups: (a) dependency annotation scheme for Indic languages based on traditional Indian Paninian grammar has been defined (Begum et al., 2008), (b) dependency annotated corpora are being created, and (c) parsers which work across Indic languages are being experimented with (Bharati et al., 2009; Bhat, 2017).

# 8 Conclusions and Future Work

## 8.1 Conclusions

We have presented an extensive case-study on translation involving 9 major Indic languages covering 72 language pairs. We experiment with the different translation units. The following are the major findings:

1. The case study provides evidence that OS and BPE-level translation models perform significantly better than word and morpheme-level models. We also observe that OS and BPE-level translation models are better than approaches relying on the transliteration of untranslated words in the output a word-level translation model.

2. The major trends we observed are: (a) subword-level translation models are more beneficial for morphologically rich Dravidian languages, (b) they are also effective when only contact relation exists between the languages.

3. Based on translation quality, we see clear partitioning of translation pairs by language family. For instance, translations involving Indo-Aryan languages can be done with a high level of accuracy, whereas those involving Dravidian languages are extremely difficult. This suggests that SMT approaches customized to language family pairs may be investigated.

4. Rich morphology of Indic languages, especially Dravidian languages, is a major factor impacting translation quality. For instance, it is easiest to translate to/from Hindi (a language with a relatively isolating morphology). On the other hand, translation involving Malayalam (a highly agglutinative language) is the most difficult.

We also report English to Indic language as well as Indic language to English translation results. We observe that word order divergence between English and Indic languages impacts translation quality. We show that English-Indic language translation quality can be improved by source-side pre-ordering. We observe that the same set of pre-ordering rules gives improvements in translation quality across all Indic languages, showing that resources developed for one language can be reused for other related languages.

The case-study thus spans 90 language pairs (90 Indic-Indic, 9 English-Indic and 9 Indic-English language pairs). These findings confirm findings from previous small-scale studies (Kunchukuttan and Bhattacharyya, 2016b, 2017) on a large number of language pairs.

### 8.2 Future Work

The following are some tasks that can be done in future to extend the case-study and, in general, improve machine translation for Indic languages:

- Subject to availability of parallel corpora, we would like to extend this study to more Indic languages. Some of the major Indic languages that have not been included in this study are Kannada, Urdu, Odia, Sindhi and Assamese. We would also like to cover major languages from the Austro-Asiatic (*e.g.* Santali, Mundari, Khasi) and the Sino-Tibetan (*e.g.* Bodo, Meitei) families.
- It would be interesting to have an exhaustive study of pivot-based SMT for Indic languages, experimenting with different pivot languages as well as combinations of pivot languages. These studies could be useful for further insights into the choice of pivot languages and relatedness among the Indic languages.
- Investigation of a common framework for addressing structural divergence in Indic language to English translation is an important research direction to pursue.
- An extensive case-study for Neural Machine Translation involving Indic languages is another relevant direction of work. In particular, it would be interesting to explore multilingual models trained at the subword-level, where lexical similarity and pooling parallel corpus resources can be tested together.